\pdfoutput=1

\documentclass[11pt,a4paper]{article}
\usepackage{makecell}
\usepackage{booktabs}
\usepackage{multirow}
\usepackage{graphicx}
\usepackage[table]{xcolor}
\usepackage{makecell}
\usepackage{caption}
\usepackage[margin=1in]{geometry}

\usepackage[table]{xcolor}
\definecolor{cvprblue}{RGB}{0,114,178} 
\usepackage[final]{acl}
\usepackage{amssymb}

\usepackage{times}
\usepackage{latexsym}
\usepackage{amsmath}
\usepackage[T1]{fontenc}

\usepackage[utf8]{inputenc}

\usepackage{microtype}

\usepackage{inconsolata}
\usepackage{algorithm}
\usepackage{algpseudocode}
\usepackage{graphicx}
\usepackage{graphicx} 
\usepackage{subfigure}
\usepackage{amsmath}
\usepackage{adjustbox}
\usepackage[]{acl}
\usepackage[T1]{fontenc}
\usepackage[utf8]{inputenc}
\usepackage{float}
\usepackage{graphicx}
\usepackage{caption}
\usepackage{subcaption}
\usepackage{calligra}
\usepackage{booktabs}
\usepackage{multirow}
\usepackage{float}
\usepackage[dvipsnames]{xcolor}
\usepackage{diagbox}
\usepackage{longtable}
\title{Towards Efficient CoT Distillation: Self-Guided Rationale Selector for Better Performance with Fewer Rationales}

\author{Jianzhi Yan\textsuperscript{1,2}, Le Liu\textsuperscript{1,2}, Youcheng Pan\textsuperscript{2}\footnotemark[1], Shiwei Chen\textsuperscript{1,2} \\ \textbf{Yang Xiang\textsuperscript{2,3}\thanks{Corresponding authors.},} \textbf{Buzhou Tang\textsuperscript{1,2}\footnotemark[1]} \\
    \textsuperscript{1}Harbin Institute of Technology, Shenzhen, China \\
    \textsuperscript{2}Pengcheng Laboratory, Shenzhen, China \\
    \textsuperscript{3}Shaoguan Research Institute of Data Industry, China \\
    \texttt{\{yanjzh, liul07, panych, chenshw ,xiangy\}@pcl.ac.cn}\\  \texttt{tangbuzhou@gmail.com} \\
    }

\begin{document}
\maketitle
\begin{abstract}

Chain-of-thought (CoT) distillation aims to enhance small language models' (SLMs) reasoning by transferring multi-step reasoning capability from the larger teacher models. However, existing work underestimates rationale quality, focusing primarily on data quantity, which may transfer noisy or incorrect information to the student model. To address the above issues, we proposed \textbf{M}odel-\textbf{O}riented \textbf{R}ationale \textbf{S}election \textbf{D}istillation (MoRSD), which can discern and select high quality rationales for distillation to improve performance further. We further propose a Rationale Difficulty (RD) metric to measure the ability of the student model to generate the correct answer under a given rationale. Compared to the baseline, we achieved 4.6$\%$ average improvement on seven datasets over three tasks, using fewer rationales by controlling their accuracy, diversity, and difficulty. Our results reveal that a small portion of the high quality rationales can enhance the reasoning ability of student models than the entire dataset. Our method promises to be a possible solution for efficient CoT distillation. Our code will be released in \url{https://github.com/Leon221220/MoRSD}.
\end{abstract}

\section{Introduction}

Large language models (LLMs) such as LLaMA, GPT-4, Gemini, DeepSeek-V3, and DeepSeek-R1, have achieved remarkable performance in various reasoning tasks by instructing them to \textit{think step-by-step} \cite{touvron2023llamaopenefficientfoundation, openai2024gpt4technicalreport, gemini_2024, deepseekai2024deepseekv3technicalreport, deepseekai2025deepseekr1incentivizingreasoningcapability, brown2020language, sun2021ernie}. Engaging in reasoning through logically coherent steps has substantially enhanced performance in tasks such as mathematical problem solving and question answering. These intermediate reasoning steps are referred to as \textit{rationale} \cite{wei2023chainofthought}.

\begin{figure}[!t]
\centering
\includegraphics[scale=0.62]{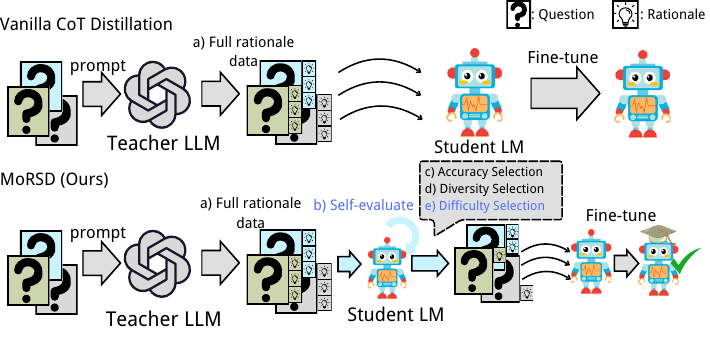}
\caption{\textbf{Vanilla CoT Distillation and MoRSD.} Different from previous studies that mostly use a), c), and d), we propose \textcolor{SkyBlue}{b)} and \textcolor{SkyBlue}{e)} to select effective data for specific student models to improve performance further.}
\label{Overview}
\end{figure}

To achieve emergent reasoning abilities, LLMs require large-scale parameters, making SLMs inherently limited \cite{wei2023chainofthought, kojima2023large, fu2023specializingsmallerlanguagemodels}. CoT distillation has become a key technique for enhancing SLM reasoning by transferring rationales from stronger teachers \cite{wang2023scott, li-etal-2023-symbolic}, showing strong results on arithmetic and symbolic tasks \cite{ho2023large, hsieh2023distilling, ying2024llmsasinstructorslearningerrorsautomating, kim2024smalllanguagemodelsequation}. Beyond basic distillation, recent works explore consistency enforcement \cite{chen2023mcckdmulticotconsistentknowledge}, cross-task supervision \cite{li-etal-2024-mode}, and tailored strategies \cite{Zhang_2024}. Mentor-KD \cite{lee-etal-2024-mentor} introduces intermediate models for better supervision, MCC-KD promotes consistent yet diverse reasoning \cite{chen2023mcckdmulticotconsistentknowledge}, while Lion \cite{kim2024smalllanguagemodelsequation} and TA-in-the-Loop \cite{Zhang_2024} use adversarial and auxiliary guidance, respectively.

However, these approaches often require additional models, discard useful failures, or introduce iterative overhead—resulting in high computational costs and limited flexibility. And many works still rely on enlarging the rationale set (increasing from 1 to 8 per instance \cite{ho2023large}) to improve performance, while \textbf{overlooking rationale quality}. Such data scaling ignores variance in correctness and diversity, risking the distillation of noisy signals. Furthermore, most approaches \textbf{neglect the specificity of student models}, failing to adapt to their strengths or limitations. These limitations motivate our focus: how to select a small set of high-quality, student-aware rationales for efficient and effective distillation.


To overcome these limitations, we propose \textbf{MoRSD}, a simple but effective method that enables student models to customize their distillation data autonomously. As presented in Figure \ref{Overview}, MoRSD consists of four stages: 1) rationale generation, 2) self-evaluation, 3) rationale selection and 4) distillation. The rationale generation stage prompts the teacher LLM to generate the rationale dataset. In the self-evaluation stage, we calculate rationale difficulty (RD) to measure the contribution of a given rationale to distillation. Specifically, RD measures the student's ability to generate the correct answer given a question and rationale. Those with smaller RD are considered more beneficial to generate the corresponding answer.


\par Then, we first apply model-agnostic accuracy selection and diversity selection to the rationale dataset. Accuracy selection adjusts the proportion of correct rationales in the dataset to achieve the given accuracy threshold, diversity selection involves pairwise Jaccard similarity to eliminate similar rationale in the dataset. Finally, we use difficulty selection to select the rationales with smaller RD. Since difficulty selection uses perplexity-based RD, a model-specific metric, it enables the student model to customize its distillation data during the difficulty selection. Through these stages, we obtain a small amount of high-quality rationale data to improve distillation performance for specific student models. In summary, our contributions are three-fold:

\par1. We propose \textbf{MoRSD}, a simple and effective method that performs better with fewer rationales. Prove that using a small portion of the dataset can outperform using the entire dataset in enhancing the reasoning ability of student models.
\par2. We propose a model-specific metric, rationale difficulty, to measure rationale contribution for distillation, enabling student models to customize data based on their training requirements.
\par3. We conducted extensive experiments on seven datasets covering three distinct tasks. The results demonstrate that our method consistently outperforms the baselines, achieving an average accuracy improvement of 4.6\%.

\section{Related work}

\subsection{Chain-of-thought (CoT) Distillation}
\par Chain-of-thought prompting delivers strong performance but typically benefits from large models with many parameters, resulting in high computational costs and limited deployment \cite{hoffmann2022an, chowdhery2022palmscalinglanguagemodeling}. \citet{ho2023large} first introduced fine-tune-CoT, a method that transfers the multi-step reasoning ability of LLMs to smaller models through fine-tuning. Some approaches use in-context learning to implicitly transfer knowledge \cite{rajani2019explain, wang2023pinto}, while others treat rationale generation as a multi-task fine-tuning objective \cite{hsieh2023distilling}. Furthermore, \citet{li-etal-2024-mode} distill the rationale into multiple experts in low-rank adaptation (LoRA), decoupling CoT reasoning from the student model. \citet{Zhang_2024} enhances knowledge transfer through active learning and explanation-guided sample selection. Some researchers identify influential tokens using gradient attribution techniques such as saliency maps to guide the student model \cite{10.1007/978-3-031-70239-6_3}. Recently, a study found that only a small fraction (4.7\%) of CoT steps are critical for performance \cite{dai2024imitationlearningkeyreasoning}, which closely matches our findings. \citet{busbridge2025distillationscalinglaws} introduce a distillation scaling law to optimize compute allocation between teacher and student models, providing efficient distillation strategies that outperform supervised pretraining in certain cases.

\begin{figure*}[ht]

	\centering
        \includegraphics[width=1.0\textwidth]{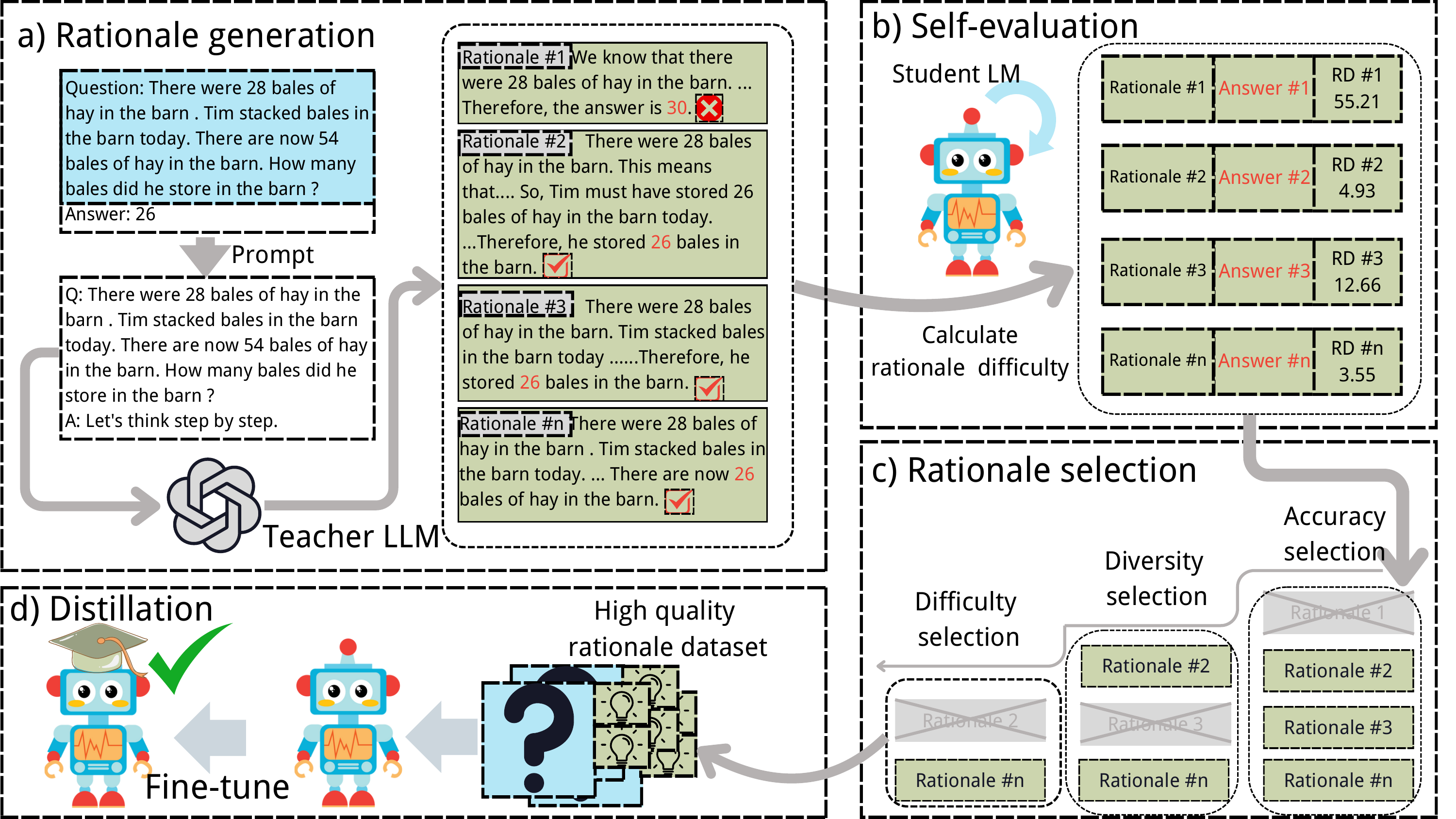} \\
	
 \caption{\textbf{Detailed overview of MoRSD}. MoRSD comprises four stages: \textbf{a) Rationale generation} prompts a teacher model to generate all the data required for the rationale selection stage (blue). \textbf{b) Self-evaluation}, which uses the rationale difficulty (RD) to evaluate all generated rationales. Those rationales with smaller RD are considered helpful for distillation. \textbf{c) Rationale selection}, which constructs the final dataset for distillation by controlling the original dataset's accuracy, diversity, and difficulty. \textbf{d) Distillation}, which fine-tunes the student model using the constructed dataset.}
\label{Detailed overview}
\end{figure*}

\subsection{Data Efficiency in Language Models}
\par Data efficiency means that the model achieves high performance with a smaller amount of training data, maximizing the value derived from limited data. \citet{yang2024qwen25mathtechnicalreportmathematical} shows that with only 1,000 carefully selected prompts and responses, models can learn to follow specific formats and generalize effectively to new tasks.  \citet{chen2024alpagasus} used GPT-3.5 to score data difficulty, and \citet{mekala2024smaller} proposed Learning Percentage (LP) for difficulty assessment, both reduced data needs for instruction tuning. LIMA achieves strong performance with few examples, generalizing well to unseen tasks and requiring minimal instruction tuning \cite{zhou2023limaalignment}. \citet{yue2024distillinginstructionfollowingabilitieslarge} uses a multi-round distillation framework with an oracle LLM to select challenging instructions for student models, reducing the need for extensive training samples. Recently, \citet{ye2025limoreasoning} proposed the "Less is More Reasoning Hypothesis" (LIMO), demonstrating that complex reasoning can be induced with few examples when the base model has pre-trained domain knowledge. \citet{muennighoff2025s1simpletesttimescaling} introduced a test-time scaling approach using a curated dataset (s1K) and budget forcing, enabling the Qwen2.5-32B-Instruct model to outperform OpenAI's o1-preview \cite{o1} on math reasoning tasks by 27$\%$ with controlled test-time compute.


\section{Method}
\subsection{Problem definition}
CoT distillation first requires prompting the teacher model to generate rationales related to the training data. Let $\mathcal{D}=\left\{\left(q_1, a_1\right),\left(q_2, a_2\right), \ldots,\left(q_N, a_N\right)\right\}$ denote the complete dataset,  where each $\left(q_i, a_i\right)$ represents a question-answer pair and the label is available. Then, a teacher model $\mathcal{T}\left(\theta^{\mathcal{T}}\right)$ (the parameter $\theta^{\mathcal{T}}$ is inaccessible) is prompted to generates $m$ distinct rationale $\left\{\hat{r}_{i}^1, \hat{r}_{i}^2, \ldots, \hat{r}_{i}^m\right\}$ where each $\hat{r}_{i}^j$ represents a separate rationale for the question $q_i$. The complete dataset with these rationales is denoted as:

\begin{equation}
\begin{aligned}
\mathcal{D}_{\text{full}} = \biggl\{ q_i, \bigl\{ & (\hat{r}_{i}^1, \hat{a}_i^1), (\hat{r}_{i}^2, \hat{a}_i^2), \ldots, (\hat{r}_{i}^j, \hat{a}_i^j) \bigr\} \biggr\}
\end{aligned}
\label{d_full}
\end{equation}

Where $i = 1, 2, \ldots, N, j = 1, 2, \ldots, M$. The performance of the student model $\mathcal{S}$ on the test set $\mathcal{D}_{\text {test }}$ can be denoted as:

\begin{equation}\label{evaluation}
    \operatorname{Perf}\left(\mathcal{S}, \mathcal{D}_{\text {test }}\right)=\frac{1}{\left|\mathcal{D}_{\text {test }}\right|} \sum_{(q, a) \in \mathcal{D}_{\text {test}}} \mathbb{I}(\mathcal{S}(q)=a)
\end{equation}

\par Our goal is to select a subset $\mathcal{D}_{\text {selected}} \subseteq \mathcal{D}_{\text {full}}$ from $\mathcal{D}_{\text {full}}$ and make the performance of the student model $\mathcal{S}_{\mathcal{D}_{\text {selected}}}$, distilled using $\mathcal{D}_{\text {selected}}$ on the test set $\mathcal{D}_{\text {test}}$, outperform that of the student model $\mathcal{S}_{\mathcal{D}_{\text {full}}}$ distilled using the full data:
\begin{equation}
\mathcal{D}_{\text {selected}}^{*}=\arg\max_{\mathcal{D}_{\text {selected}} \subseteq \mathcal{D}_{\text {full}}} \operatorname{Perf}\left(\mathcal{S}_{\mathcal{D}_{\text {selected }}}, \mathcal{D}_{\text {test }}\right)
\end{equation}


To achieve the above goal, we designed a four-stage distillation method MoRSD. Its details will be described in the following sections.

\subsection{Rationale generation}\label{Rationale Generation}
To obtain the dataset for distillation, we adopt the same generation method as in previous studies\cite{ho2023large}. As shown in the upper left of Figure \ref{Detailed overview}, we use a fixed template: "Q: $\left\langle q_i\right\rangle$. A: Let’s think step by step. \textcolor{blue}{$\left\langle\hat{r}_i\right\rangle$} Therefore, the answer is \textcolor{red}{$\left\langle\hat{a}_i\right\rangle$}". By applying this process to all data points in $\mathcal{D}$, we obtain the full dataset $\mathcal{D}_{\text {full}}$ in Eq \ref{d_full}.

\subsection{Self-evaluation}\label{self-evaluation}
After building the full dataset $\mathcal{D}_{\text {full}}$ in Section \ref{Rationale Generation}, we use rationale difficulty (RD) to score each rationale $r_{i}^j$ in the dataset. RD is a metric based on the perplexity of the student model, where perplexity is the exponential transformation of the normalized Negative Log-Likelihood (NLL), given an input sequence $X=\left(x_1, x_2, \ldots, x_N\right)$ and a target sequence $Y=\left(y_1, y_2, \ldots, y_M\right)$, the perplexity can be written as:

{\small
\begin{equation}
\text{PPL}(y_j|X) = \exp\left(
-\frac{1}{M}\sum_{j=1}^{M}\log \operatorname{Pr}(y_j|x_1, ..., x_N, y_{j-1})\right) 
\end{equation}
}

Since the student model has been pre-trained or supervised-fine-tuned (SFT) using NLL loss on a large corpus of text, its perplexity can indicate the quality of the rationales generated by the teacher. Therefore, we define RD as the ratio of the change in PPL of the student model before and after a given rationale:

\begin{equation}
    RD(\hat{r}_i^j, q_i) = \frac{\text{PPL}_{(\theta^{\mathcal{S}})}(a_i | \hat{r}_i^j, q_i)}{\text{PPL}_{(\theta^{\mathcal{S}})}(a_i | q_i)}.
\label{eq1}
\end{equation}

For rationale $\hat{r}_i^j$, if the student model achieves low $RD(\hat{r}_i^j, q_i)$, it suggests that the rationale is more beneficial for the student in understanding the corresponding question and will be selected in difficulty selection.

\subsection{Rationale selection}\label{Rationale Selection}
After calculating the RD for each rationale in section \ref{self-evaluation}, this section will select a subset $\mathcal{D}_{\text {selected}}$ from the full dataset $\mathcal{D}_{\text {full}}$ based on the accuracy, diversity, and difficulty of the rationale. Therefore, we divide the rationale selection process into three sequential parts: 1) Accuracy Selection, 2) Diversity Selection, and 3) Difficulty Selection.

\subsubsection{Accuracy selection}\label{acc selection}
The most important characteristic of rationale is correctness. Different from \cite{ho2023large, li-etal-2024-mode}, we first divide the rationale into correct and incorrect parts by comparing the final prediction $\hat{a}_i$ of the teacher model with the ground truth $a_i$. We then filter out negative samples to ensure the original dataset meets a given accuracy threshold $\delta$. 

Then, we filter the rationales sequentially from the original dataset such that the average accuracy of the filtered dataset $\mathcal{D}_{\text {accurate}}$ reaches $\delta$. The calculation is as follows:

{\small
$$
\operatorname{Avg} \operatorname{Acc}=\frac{1}{\left|\mathcal{D}_{\mathrm{accurate}}\right|} \sum_{\left(\hat{r}_i^j, \hat{a}_i\right) \in \mathcal{D}_{\mathrm{accurate}}} \operatorname{acc}\left(\hat{r}_i^j, \hat{a}_i\right) \geq \delta
$$
}

\begin{table*}[t]
\centering
\resizebox{\textwidth}{!}{
\begin{tabular}{lc|llllllll}
\toprule
\multirow{2}{*}{\textbf{Method}} & \multirow{2}{*}{\textbf{Params}} 
& \textbf{Single} & \textbf{Add} & \textbf{Multi} & \textbf{Strategy} 
& \multirow{2}{*}{\textbf{GSM8K}} & \multirow{2}{*}{\textbf{SVAMP}} & \textbf{Date} & \textbf{Shuffled} \\
                                &                                  & \textbf{Eq} & \textbf{Sub} & \textbf{Arith} & \textbf{QA} &  &  & \textbf{Understanding} & \textbf{Objects} \\
\midrule
\textbf{Random}               & -                              & 0.00 & 0.00 & 0.00 & 50.00 & 0.00 & 0.00 & 17.12 & 33.33 \\
\midrule
\multicolumn{10}{c}{\textbf{Teacher:  InstructGPT 175B (text-davinci-002)}} \\
\midrule
\multirow{1}{*}{\textsc{Zero-shot-CoT}} & 175B & 82.24 & 78.99 & 78.89 & 53.57 & 40.26 & 64.67 & 73.87 & 50.22 \\
\midrule
\multicolumn{10}{c}{\textbf{Student: Flan-T5-$\{$Small, Base, Large, XL$\}$}} \\
\midrule
\multirow{4}{*}{\makecell[l]{\textsc{Vanilla}\\\textsc{CoT Distillation}}}
                                      & 60M    & 7.24  & \colorbox{cvprblue!14}{\textbf{10.92}} & 17.22 & 56.04 & 2.58 & 10.67 & \colorbox{cvprblue!14}{\textbf{84.68}} & 62.22 \\
                                      & 250M   & \colorbox{cvprblue!14}{\textbf{9.21}} & 10.92 & 21.11 & 60.84 & 4.40 & 12.33 & 84.68 & 67.11 \\
                                      & 780M   & 10.52 & 15.13 & 20.00 & 61.72 & 7.12 & 13.67 & 87.39 & 89.33 \\
                                      & 3B     & 20.39   & 11.76   & 26.67   & 65.37   & 7.60 & 12.33   & 82.9   & 43.11 \\
\midrule
\multirow{3}{*}{\makecell[l]{\textsc{Multi-task}\\\textsc{CoT Distillation}}}
                                      & 250M   & 5.22  & 8.40  & 8.33  & 52.83 & 6.00 & 2.33  & 80.18 & 31.55 \\
                                      & 780M   & 11.89 & 16.81 & 16.81 & 50.09 & 6.36 & 9.00  & 79.23 & 35.96 \\
                                      & 3B     & 22.36   & \colorbox{cvprblue!14}{\textbf{36.9}}   & 17.22   & 52.11   & 7.73 & 11.33   & 81.93   & 52.46 \\
\midrule
\multirow{3}{*}{\textsc{MoDE-CoTD}} 
                                      & 250M   & 5.26  & 7.56  & 13.89 & 56.18 & 6.11 & 5.33  & 85.55 & 35.55 \\
                                      & 780M   & 10.52 & 10.92 & 13.89 & 56.47 & 7.28 & 11.33 & \colorbox{cvprblue!14}{\textbf{89.19}} & 62.22 \\
                                      & 3B     & \colorbox{cvprblue!14}{\textbf{23.33}}   & 24.37   & 23.33   & 60.99   & 9.78 & 17.33   & \colorbox{cvprblue!14}{\textbf{93.69}} & 70.67 \\
\midrule
\multirow{4}{*}{\textsc{MoRSD (OURS)}} 
                                      & 60M    & \colorbox{cvprblue!14}{\textbf{9.21}} & \colorbox{cvprblue!14}{\textbf{10.92}} & \colorbox{cvprblue!14}{\textbf{22.78}} & \colorbox{cvprblue!14}{\textbf{60.26}} & \colorbox{cvprblue!14}{\textbf{6.98}} & \colorbox{cvprblue!14}{\textbf{11.33}} & 82.88 & \colorbox{cvprblue!14}{\textbf{83.56}} \\
                                      & 250M   & \colorbox{cvprblue!14}{\textbf{9.21}} & \colorbox{cvprblue!14}{\textbf{12.61}} & \colorbox{cvprblue!14}{\textbf{24.44}} & \colorbox{cvprblue!14}{\textbf{65.65}} & \colorbox{cvprblue!14}{\textbf{6.98}} & \colorbox{cvprblue!14}{\textbf{13.67}} & \colorbox{cvprblue!14}{\textbf{86.49}} & \colorbox{cvprblue!14}{\textbf{99.56}} \\
                                      & 780M   & \colorbox{cvprblue!14}{\textbf{13.16}} & \colorbox{cvprblue!14}{\textbf{16.81}} & \colorbox{cvprblue!14}{\textbf{25.00}} & \colorbox{cvprblue!14}{\textbf{65.65}} & \colorbox{cvprblue!14}{\textbf{9.71}} & \colorbox{cvprblue!14}{\textbf{15.00}} & \colorbox{cvprblue!14}{\textbf{89.19}} & \colorbox{cvprblue!14}{\textbf{100.00}} \\
                                      & 3B     & 21.71   & 24.37   & \colorbox{cvprblue!14}{\textbf{31.67}}   & \colorbox{cvprblue!14}{\textbf{65.65}}   & \colorbox{cvprblue!14}{\textbf{10.20}} & \colorbox{cvprblue!14}{\textbf{23.67}}   & 91.00   & \colorbox{cvprblue!14}{\textbf{100.00}} \\
\bottomrule
\end{tabular}
}
\caption{\textbf{MoRSD Performance}. Accuracy ($\%$) of MoRSD and baseline methods on 8 tasks under various settings. \textbf{Random} refers to random-guess performance derived based on the number of choices in multi-choice tasks. The best method for each setting is marked in \colorbox{cvprblue!14}{\textbf{bold}}. For \textbf{Zero-shot-CoT}, we use the same prompt setting as \cite{ho2023large}.}
\label{main result}
\end{table*}

\subsubsection{Diversity selection}
The diversity of rationales is important for distillation performance. However, we found that even with different sampling temperatures, the teacher model often generates similar rationales. To address this, we select diverse rationales by first splitting them into N-grams (N=3 in our experiments). Then, we calculate the pairwise Jaccard similarity between these N-gram sets. For each rationale $r_i^j$, we decompose it into segments $R_i^j$ and use the Jaccard similarity score to compare and identify the most similar rationales.

\begin{equation} 
(r_i^m, r_i^n)=\underset{1 \leq m, n \leq M, m \neq n}{\arg \max } \frac{\left|R_i^m \cap R_i^n\right|}{\left|R_i^m \cup R_i^n\right|}
\label{eq:diversity}
\end{equation}

We then randomly keep one form the two rationales from Eq. \ref{eq:diversity} and discard the other. This process repeats until we collect a total of $K$ rationales (set to 6 in our experiments). Afterward, we have a diverse dataset, $\mathcal{D}_{\text {diverse}}$, ready for the final difficulty selection step.

\subsubsection{Diffculty selection}
After obtaining $\mathcal{D}_{\text {diverse}}$, we need to filter and retain rationales that are helpful for distillation based on RD. As mentioned in section \ref{self-evaluation}, rationale with low RD is considered helpful for distillation, so in the difficulty selection, we select the $k$ ($k$ set to 3 in our experiments) samples with the lowest RD in the dataset:

\begin{table*}[t]
\centering
\renewcommand{\arraystretch}{1.15}
\setlength{\tabcolsep}{5pt}
\caption{\textbf{Performance of \textsc{MoRSD} and baselines across two student models on four tasks.} Best results for each student model are in \colorbox{cvprblue!14}{\textbf{bold}}.}
\vspace{2pt}
\resizebox{\linewidth}{!}{
\begin{tabular}{ll|ccccc}
\toprule
\textbf{Method} & \textbf{Student} 
& \makecell[c]{\textbf{Strategy}\\\textbf{QA}} 
& \textbf{SVAMP} 
& \makecell[c]{\textbf{Date}\\\textbf{Understanding}} 
& \makecell[c]{\textbf{Shuffled}\\\textbf{Object}} 
& \textbf{Average} \\
\midrule
\textsc{MCC-KD}       & \textbf{FlanT5-Small} & 58.37 & 10.00 & 81.98 & 43.11 & 48.37 \\
\textsc{Mentor-KD}    & \textbf{FlanT5-Small} & 59.97 & 10.67 & 83.78 & 82.67 & 59.27 \\
\textsc{MoRSD (Ours)} & \textbf{FlanT5-Small} & \colorbox{cvprblue!14}{\textbf{61.35}} & \colorbox{cvprblue!14}{\textbf{12.33}} & \colorbox{cvprblue!14}{\textbf{84.43}} & \colorbox{cvprblue!14}{\textbf{84.69}} & \colorbox{cvprblue!14}{\textbf{60.70}} \\
\midrule
\textsc{MCC-KD}       & \textbf{FlanT5-Base} & 64.92 & 12.00 & 85.59 & 69.78 & 58.07 \\
\textsc{Mentor-KD}    & \textbf{FlanT5-Base} & 65.21 & 11.33 & \colorbox{cvprblue!14}{\textbf{87.39}} & 93.78 & 64.43 \\
\textsc{MoRSD (Ours)} & \textbf{FlanT5-Base} & \colorbox{cvprblue!14}{\textbf{65.72}} & \colorbox{cvprblue!14}{\textbf{14.28}} & 87.04 & \colorbox{cvprblue!14}{\textbf{99.62}} & \colorbox{cvprblue!14}{\textbf{66.67}} \\
\bottomrule
\end{tabular}
}
\label{main result2}
\vspace{-4pt}
\end{table*}

{\small
\begin{equation}
\begin{aligned}
\mathcal{D}_{\text{selected}} = \biggl\{ q_i, \bigl\{ & (\hat{r}_{i}^1, \hat{a}_i^1), (\hat{r}_{i}^2, \hat{a}_i^2), 
& \ldots, (\hat{r}_{i}^k, \hat{a}_i^k) \bigr\} \biggr\}
\end{aligned}
\label{d_selected}
\end{equation}
}

where $RD\left(\hat{r}_{i}^1, q_i\right) \leq RD\left(\hat{r}_{i}^2, q_i\right) \leq \cdots \leq RD\left(\hat{r}_{i}^k, q_i\right)$, $i = 1, 2, \ldots, N^*, j = 1, 2, \ldots, M^*$.

\subsection{Distillation}
Then, we use $\mathcal{D}_{\text selected}$ to fine-tune the student model. Similar to SFT, the objective function of distillation can be written as follows:

{\small
\begin{equation}
\mathcal{L}(\theta_{\mathcal{S}})=-\sum_{r_i \in \mathcal{D}_{\text {selected}}} \mathbf{1}_{\left(r_i\right)} \cdot \log \operatorname{Pr}\left(a_i, \hat{r_i} \mid q_i ; \theta_{\mathcal{S}}\right)
\end{equation}
}

The final distilled student model $\mathcal{D}_{\text {selected}}$ is used to verify the final performance according to Eq \ref{evaluation}.

\section{Experiment}\label{Experiment}

\subsection{Task and Datasets}
Experiments were conducted on seven datasets related to three tasks: mathematical reasoning, question answering, and temporal/spatial reasoning. Including StrategyQA \cite{geva2021did} for commonsense reasoning,  Addsub \cite{hosseini-etal-2014-learning}, Multiarith \cite{roy-roth-2015-solving}, SVAMP \cite{patel-etal-2021-nlp}, SingleEq \cite{koncel-kedziorski-etal-2015-parsing} and GSM8K \cite{cobbe2021trainingverifierssolvemath} for arithmetic math inference and Date Understanding \cite{srivastava2023imitationgamequantifyingextrapolating}, Tracking Shuffled Objects \cite{srivastava2023imitationgamequantifyingextrapolating} for temporal/spatial reasoning. The details on partition training, testing sets, and other specificities are provided in the Appendix \ref{Appendix}.

\subsection{Baseline}
We provide a comparison of MoRSD (ours) with three baseline methods:
\par $\bullet$ \textbf{Vanilla CoT Distillation} \cite{ho2023large}, where the student model is directly fine-tuned on the teacher-generated CoT rationales without additional selection or filtering.
\par $\bullet$ \textbf{Multi-task CoT Distillation} \cite{li-etal-2024-mode}, where the student model is fine-tuned on a combined dataset from multiple reasoning tasks.
\par $\bullet$ \textbf{MoDE-CoTD} \cite{li-etal-2024-mode}, where the rationales from different tasks are distilled into separate LoRA modules, enabling cross-task collaboration through task-specific parameter adaptation.
\par $\bullet$ \textbf{MCC-KD} \cite{chen2023mcckdmulticotconsistentknowledge}, which improves reasoning consistency by generating multiple rationales per question and minimizing bidirectional KL-divergence between their answer distributions.
\par $\bullet$ \textbf{Mentor-KD} \cite{lee-etal-2024-mentor}, which uses a task-specific mentor model to enrich the distillation set with CoT annotations and soft labels, addressing data quality and label scarcity.

\begin{figure}[t]
	\centering 
        \includegraphics[width=0.47\textwidth]{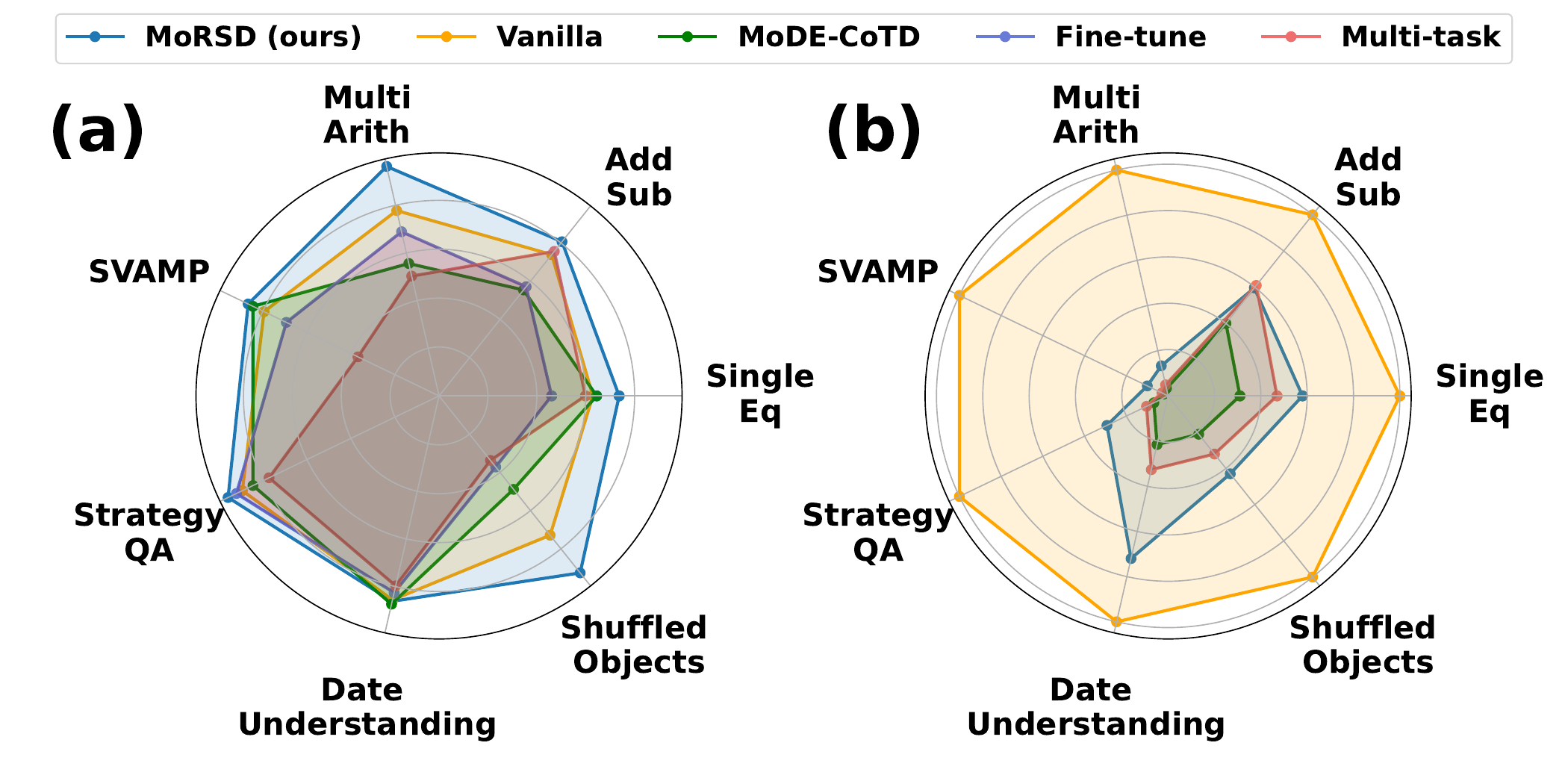} \\
 \caption{\textbf{Comparison of the performance and the rationale usage}.}
\label{data_amount_vis_radar}
\end{figure}

\subsection{Teacher and Student Models}
\par For the teacher models, we use GPT-3 175B \cite{brown2020language}, accessed via the OpenAI API, with \textit{text-davinci-002} \cite{ouyang2022traininglanguagemodelsfollow} as the default model unless otherwise specified. We employ the instruction-tuned versions of T5 for the student models, specifically Flan-T5-$\{$Small, Base, Large$\}$ \cite{chung2022scaling}.

\section{Results}

In this section, we report the performance of our MoRSD and baseline methods on 7 benchmarks. We compare our approach with baselines of different model sizes. The performance on the test set demonstrates the effectiveness of our approach, showing that our method achieves better performance with fewer samples.

\subsection{MoRSD outperforms baselines across different student models}

\begin{figure}[t]
	\centering
        \includegraphics[width=0.45\textwidth]{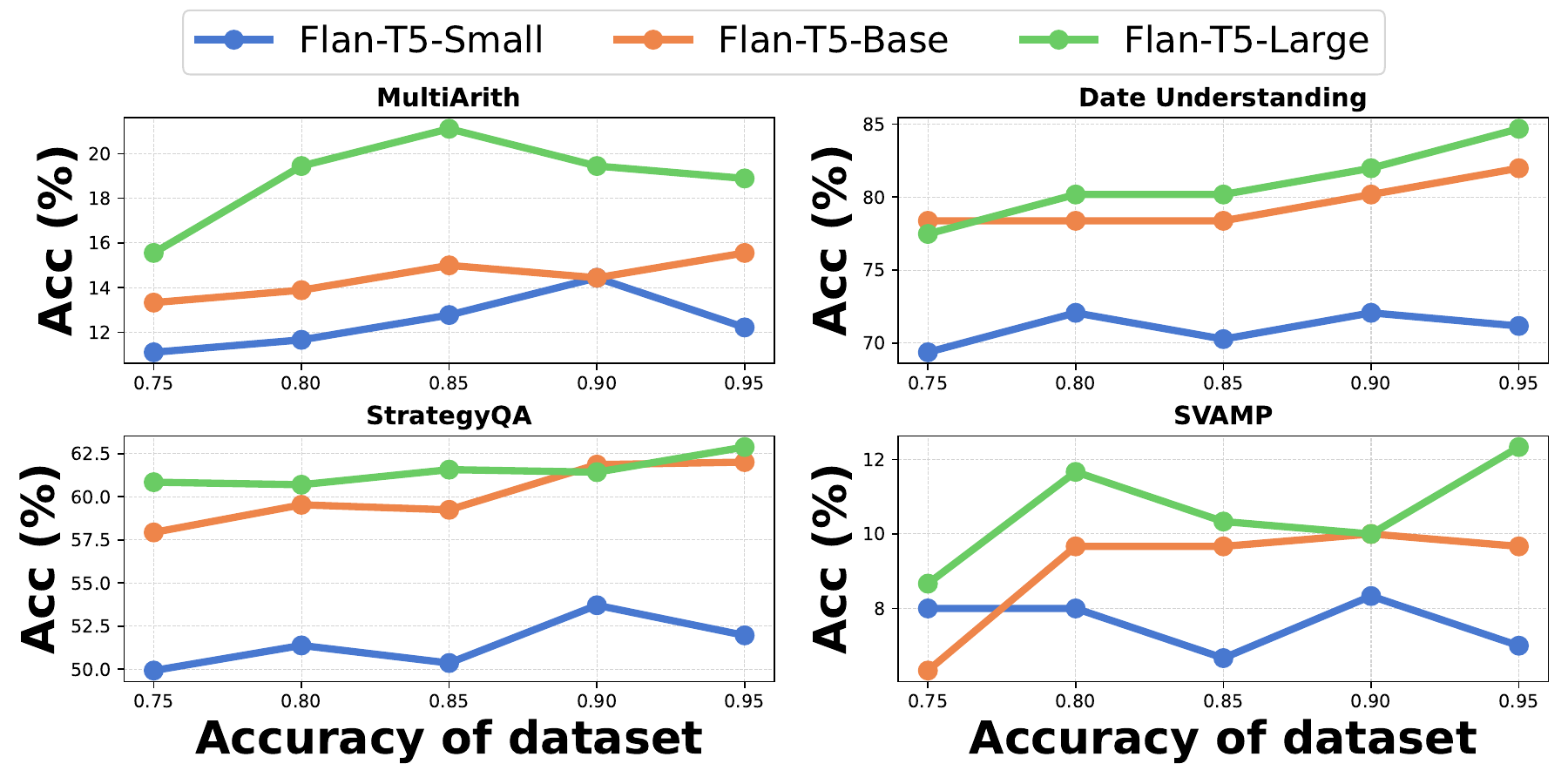} \\
 \caption{\textbf{Effect of dataset accuracy}.  The performances of MoRSD on the MultiArith, Date Understanding, StrategyQA and SVAMP datasets with different correctness rates of the teacher generated rationales.}
\label{accuracy_analyse}
\end{figure}

\begin{figure}[t]

	\centering
        \includegraphics[width=0.45\textwidth]{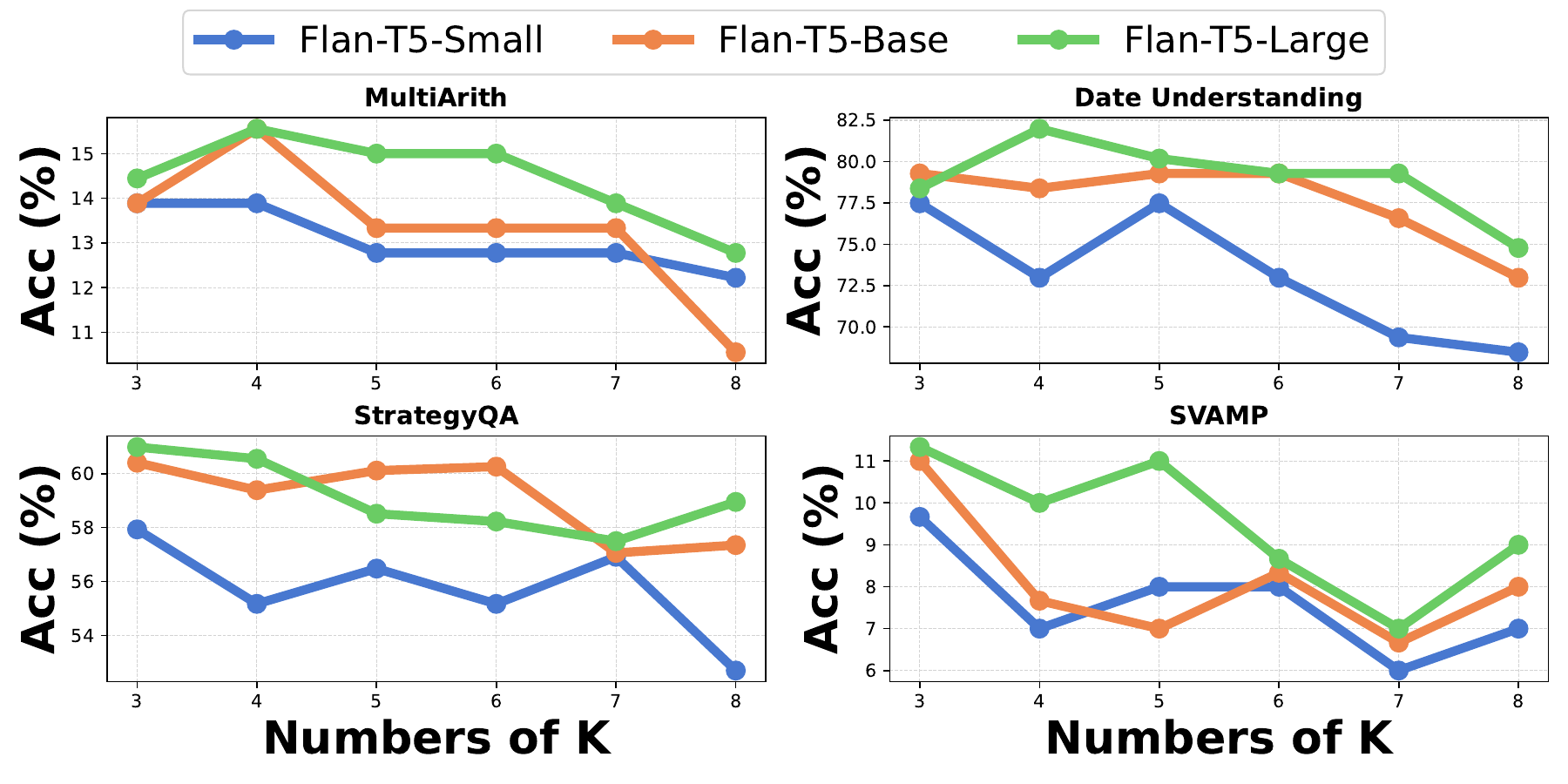} \\
	
 \caption{\textbf{Effect of rationale diversity}. The performance of MoRSD on four test sets with different rationale diversities.}
\label{diversity_analyse}
\end{figure}

The results in Table~\ref{main result} and Table~\ref{main result2} show that \textbf{MoRSD consistently outperforms strong baselines} across various student model sizes and reasoning tasks. On \textbf{Flan-T5-Small}, MoRSD notably improves results on challenging datasets such as \textbf{SVAMP} and \textbf{Tracking Shuffled Objects}, achieving \textbf{11.33\%} on SVAMP (+3.73\% over MoDE-CoTD) and \textbf{83.56\%} on Tracking Shuffled Objects, surpassing MoDE-CoTD (62.22\%) and Multi-task CoT (31.55\%). These improvements are obtained with \textbf{fewer rationales}, highlighting the effectiveness of selective rationale filtering over data quantity.

Compared to \textbf{multi-task and consistency-based methods} like \textbf{MCC-KD} and \textbf{Mentor-KD}, MoRSD achieves \textbf{comparable or better performance}. On Flan-T5-Small, it reaches an \textbf{average accuracy of 59.51\%}, slightly above Mentor-KD (59.27\%) and notably higher than MCC-KD (48.37\%), demonstrating that \textbf{effective rationale selection can boost performance without extra supervision}.

As the \textbf{student model scales up}, MoRSD \textbf{continues to outperform baselines}. On \textbf{Flan-T5-Base}, it achieves the \textbf{highest average accuracy of 66.34\%}, exceeding Mentor-KD (64.43\%) and MCC-KD (58.07\%). Notably, MoRSD achieves \textbf{near-perfect accuracy} on temporal and spatial reasoning tasks such as \textbf{Tracking Shuffled Objects (99.56\%)} and \textbf{Date Understanding (86.49\%)}, indicating strong generalization.

\subsection{Effect of rationale correctness and diversity}
\par To assess how rationale accuracy affects distillation, we varied dataset accuracy and measured student performance. As shown in Figure~\ref{accuracy_analyse}, distillation improves with higher accuracy, but gains plateau beyond a certain threshold. This indicates that accuracy is crucial at lower levels, while its marginal benefit diminishes as it increases.

The diversity of the rationale is also vital for distillation. To measure the degree of diversity among rationales, we use the number of rationales remaining after the Jaccard similarity filtering to measure the diversity of the dataset. In simple terms, a smaller number of remaining rationales after filtering indicates a higher level of diversity in the dataset. As illustrated in Figure \ref{diversity_analyse}, the performance of MoRSD exhibits a corresponding improvement with increasing diversity among the rationales, as observed in all four different test sets.


\begin{figure}[t]

	\centering
        \includegraphics[width=0.45\textwidth]{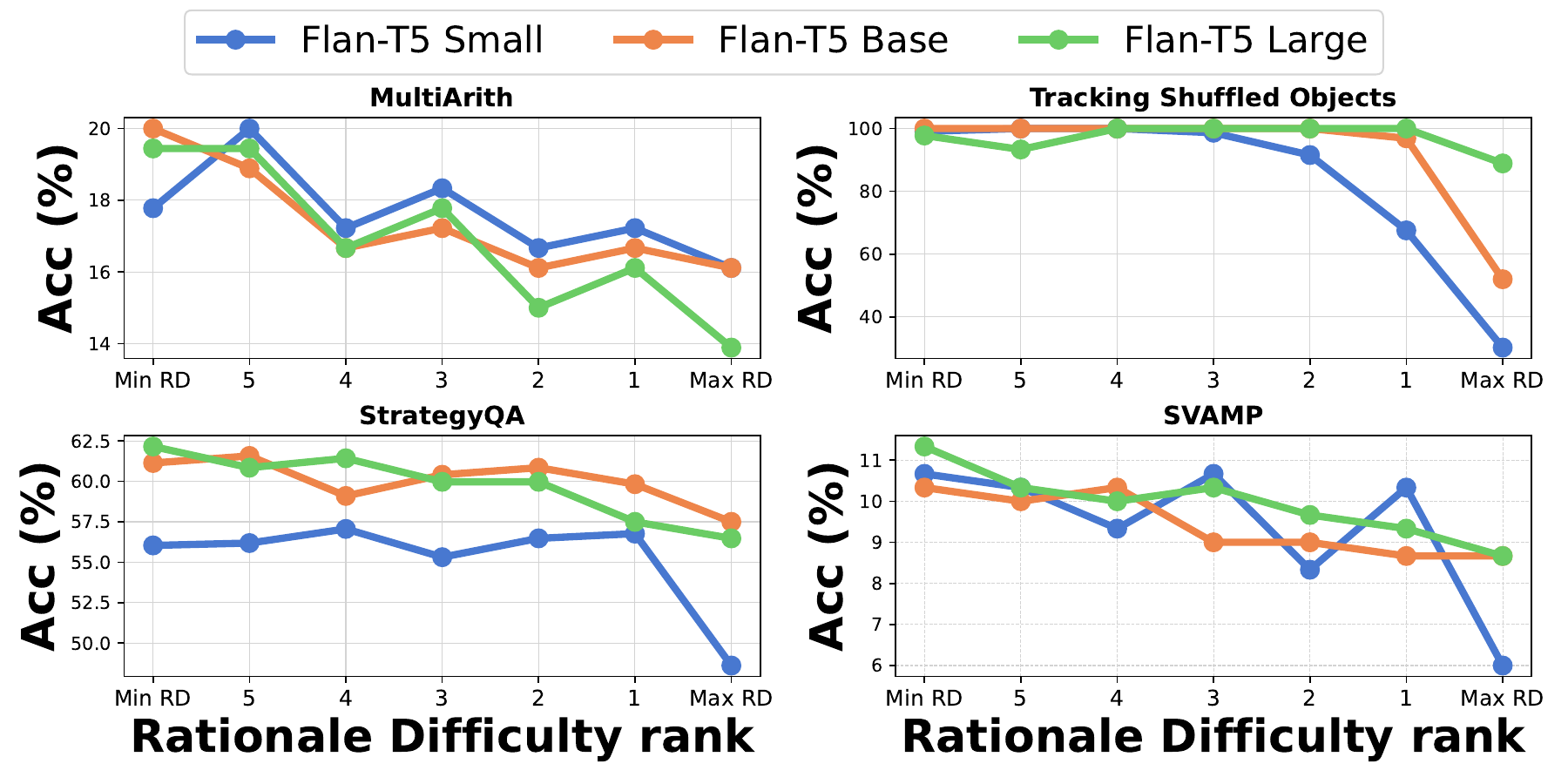} \\
	
 \caption{\textbf{Effect of rationale difficulty}.The performance of MoRSD using different samples selected by RD among four test sets}
\label{rd_rank_analyse}
\end{figure}

\begin{table*}[ht]
\centering
\resizebox{\textwidth}{!}{
\begin{tabular}{l|ccccccc}
\toprule
\textbf{Method} & \makecell[c]{\textbf{Single}\\\textbf{Eq}} & \makecell[c]{\textbf{Add}\\\textbf{Sub}} & \makecell[c]{\textbf{Multi}\\\textbf{Arith}} & \makecell[c]{\textbf{Strategy}\\\textbf{QA}} & \textbf{SVAMP} & \makecell[c]{\textbf{Date}\\\textbf{Understanding}} & \makecell[c]{\textbf{Shuffled}\\\textbf{Objects}} \\
\midrule
\textsc{MoRSD}                     & 9.21 & 10.92 & 22.78 & 60.26 & 11.33 & 82.88 & 82.22 \\
w/o \textsc{Accuracy Sel.}                 & $\text{5.92}_{\text{ \color{red}-3.29}}$ & $\text{10.08}_{\text{ \color{red}-0.84}}$ & $\text{15.00}_{\text{ \color{red}-7.78}}$ & $\text{57.21}_{\text{ \color{red}-3.05}}$ & $\text{5.67}_{\text{ \color{red}-5.66}}$ & $\text{74.77}_{\text{ \color{red}-8.11}}$ & $\text{89.33}_{\text{ \color{green}+7.11}}$ \\
w/o \textsc{Diversity Sel.}               & $\text{9.21}_{\text{ -0.00}}$ & $\text{10.92}_{\text{ -0.00}}$ & $\text{15.00}_{\text{ \color{red}-7.78}}$ & $\text{59.64}_{\text{ \color{red}-0.62}}$ & $\text{4.67}_{\text{ \color{red}-6.66}}$ & $\text{82.88}_{\text{ -0.00}}$ & $\text{67.56}_{\text{ \color{red}-14.66}}$ \\
w/o \textsc{Difficulty Sel.}             & $\text{1.97}_{\text{ \color{red}-7.24}}$ & $\text{8.40}_{\text{ \color{red}-2.52}}$ & $\text{15.56}_{\text{ \color{red}-7.22}}$ & $\text{60.26}_{\text{ -0.00}}$ & $\text{7.33}_{\text{ \color{red}-4.00}}$ & $\text{76.58}_{\text{ \color{red}-6.30}}$ & $\text{82.22}_{\text{ -0.00}}$ \\
\bottomrule
\end{tabular}
}
\caption{\textbf{Ablation study on Flan-T5-Small}. Results of ablation study about Accuracy selection, Diversity selection, and Difficulty selection on test sets.}
\label{ablation}
\end{table*}

\subsection{Effect of rationale difficulty}
To verify the effect of the rationale difficulty (RD) on distillation performance, we conducted experiments using samples of varying sizes selected after sorting based on RD. As illustrated in Figure \ref{rd_rank_analyse}, the distillation performance of the student model improves as the RD of the selected data decreases, achieving optimal performance when the RD is at its smallest. This trend is consistent across multiple test sets, including StrategyQA and Tracking Shuffled Objects, demonstrating that lower RD values correlate with more effective distillation outcomes. The results underscore the efficacy of the proposed RD indicator in identifying and prioritizing data that is most beneficial for the distillation process. This finding highlights the importance of RD in enhancing the overall performance of the student model by focusing on the most informative and manageable rationales.

\subsection{Ablation study}
\par In this section, we conduct an ablation study on the Flan-T5-Small model to assess the contributions of accuracy, diversity, and difficulty selection in MoRSD. As shown in Table~\ref{ablation}, removing any component leads to notable performance drops. Accuracy selection is critical—its removal causes large degradations on tasks like \textbf{SingleEq} ($-35.7\%$) and \textbf{SVAMP} ($-38.1\%$). Diversity selection is especially important for reasoning-heavy tasks such as \textbf{MultiArith} ($-31.2\%$) and \textbf{Tracking Shuffled Objects} ($-44.3\%$), helping reduce redundancy. Difficulty selection prioritizes informative rationales, and its absence also leads to significant drops, including $-44.0\%$ on \textbf{SingleEq} and $-26.9\%$ on \textbf{SVAMP}. These results indicate that each selection stage plays a distinct and complementary role in improving distillation effectiveness. Overall, all three components are essential for maximizing student performance.

\subsection{Analyse of selected rationale}\label{Analyse of selected rationale}

\begin{figure}[ht!]

	\centering
        \includegraphics[width=0.45\textwidth]{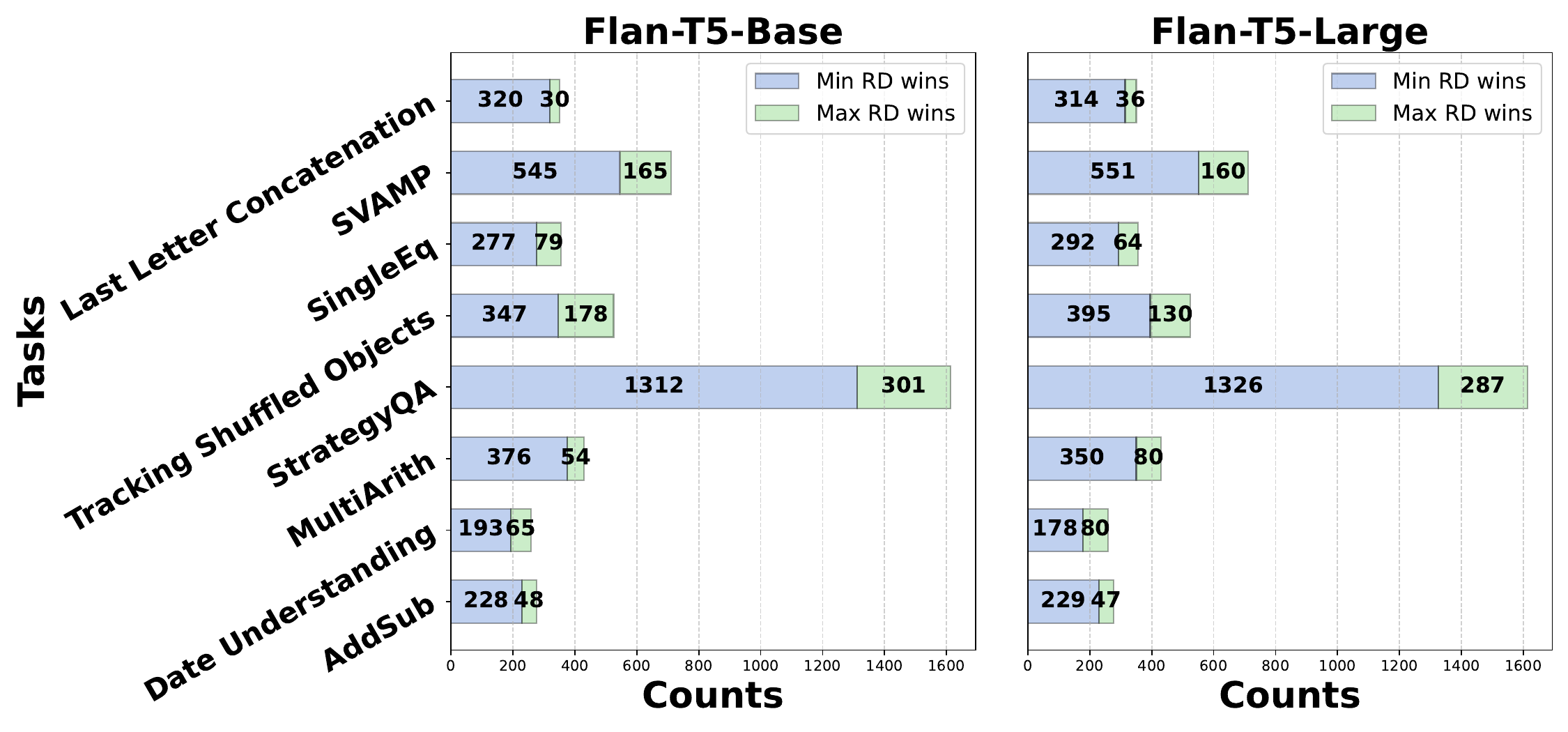} \\
	
 \caption{\textbf{Effect of selected rationale}. The ChatGPT API was used as a referee, prompted to compare two rationales and rate them on a scale of 1 to 10. Each rationale pair (maximum RD and minimum RD) was judged twice to avoid position bias, with the rationale positions swapped in each evaluation.}
\label{rationale_analyse}
\end{figure}

In order to compare the quality of rationales screened by different methods, we introduced the ChatGPT API as a referee to further explore the characteristics of different rationales selected using RD. By stitching different rationales together and prompting the referee to judge which of the two is better and give them a score of 1-10, we visualized these results as the winning frequency of those selected with the minimum RD and the maximum RD. As presented in Figure \ref{rationale_analyse}, to avoid possible bias of the judges due to the position of the rationale in the prompt, we judged each maximum RD-minimum RD pair twice and exchanged the position of the rationale in the prompt in each judgment. From the results, we can conclude that the quality of rationales with lower RD attributes is higher than those with higher RD attributes on all datasets. This further proves the effectiveness of the RD in selecting high-quality rationales.

\begin{table*}[ht]
\centering
\begin{tabular}{lccccc}
\toprule
\textbf{Method Setting} & \textbf{Label Type} & \makecell{SVAMP} & \makecell{Strategy\\QA} & \makecell{Date\\Understanding} & \makecell{Shuffled \\ Objects} \\
\midrule
Positive-only & Ground-truth  & 13.2 & 64.9 & 85.2 & 98.4 \\
MoRSD & Ground-truth  & 13.7 & 65.7 & 86.5 & 99.6 \\
MoRSD + Predict label & Teacher prediction & 10.2 & 61.7 & 82.0 & 96.1 \\
\bottomrule
\end{tabular}
\caption{Ablation study on labeling strategies in MoRSD (Flan-T5-Base). 
\textbf{Positive-only} uses only gold rationales and answers, 
\textbf{MoRSD} applies RD-based rationale selection with gold supervision and achieves the best overall accuracy, 
while \textbf{MoRSD + Predict label} replaces gold labels with teacher predictions and suffers clear degradation. 
Results show that RD-based selection improves data efficiency, but gold answer supervision is crucial.}

\label{tab:method_settings}
\end{table*}

\subsection{Effect of negative rationale}
To further examine the role of imperfect rationales and supervision signals, we conduct an ablation study across three labeling strategies, as summarized in Table~\ref{tab:method_settings}. The Positive-only setting relies exclusively on gold rationales and answers, achieving reasonable performance but with limited diversity. Our default MoRSD configuration, which selects rationales based on Rationale Difficulty (RD) while always supervising with the gold answer, yields consistent improvements across all tasks (e.g., +0.8 on StrategyQA and +1.3 on Date Understanding) and achieves the best overall average (66.4). This confirms that RD-based selection enhances both quality and diversity of rationales without sacrificing correctness. In contrast, replacing gold answers with teacher-predicted labels substantially degrades performance (e.g., a 3.9 drop on SVAMP and 4.0 on Date Understanding), highlighting the necessity of grounding training supervision in correct labels. These results validate our design choice: selectively retaining imperfect rationales improves robustness and data efficiency, but correctness of the final answer supervision remains critical.

\section{Discussion}
\subsection{Inclusion of Negative Rationales}
An important design decision in MoRSD concerns the treatment of rationales that do not lead to correct teacher predictions. While conventional approaches often discard such negative rationales entirely, we deliberately retain a subset that passes our Rationale Difficulty (RD) filter. Specifically, a rationale is preserved if it reduces the student’s perplexity in predicting the ground-truth answer, even when the intermediate reasoning is partially incorrect. This choice is motivated by prior findings that structural patterns in CoT traces can facilitate learning even when their semantic content is imperfect. 

\subsection{Generalization Across Tasks and Domains}
While MoRSD demonstrates consistent improvements on diverse reasoning benchmarks, its current evaluation scope remains primarily within math and structured reasoning tasks. An open question is how well the rationale selection paradigm generalizes to other domains, where the structure of rationales may differ significantly from mathematical derivations. Moreover, in multilingual or cross-domain scenarios, the reliability of perplexity-based Rationale Difficulty (RD) as a selection signal could be weakened, since student models may not share the same linguistic or distributional priors as their teachers. Future work could explore extending MoRSD to multilingual reasoning tasks, domain-adaptation settings, thereby testing whether the principle of less but better rationales remains universally effective beyond the current experimental scope.

\subsection{Considerations for RD-Based Selection}
In MoRSD, Rationale Difficulty (RD) serves as the central criterion for rationale selection by measuring the student’s perplexity reduction on gold answers. While effective, RD captures only part of rationale quality: it reflects token-level uncertainty but may not align with logical soundness or pedagogical value. In addition, distributional mismatch across domains or languages could further reduce its reliability. Future work could extend MoRSD by integrating RD with complementary signals—such as process reward models or structural coherence metrics—to achieve more robust rationale evaluation.

\section{Conclusion}
In this work, we propose \textbf{MoRSD}, an efficient CoT distillation method that enhances the performance of small language models using fewer rationales. By introducing a self-guided Rationale Difficulty metric, MoRSD enables the autonomous selection of high-quality rationales, effectively addressing challenges related to the rationale quality. Experiments across seven datasets demonstrate an average accuracy improvement of 4.6$\%$ over the baseline. MoRSD outperforms full dataset distillation with a small, tailored set of rationales, providing a robust solution for efficient CoT distillation and advancing knowledge transfer in a more efficient manner.

\section*{Limitations}

Although MoRSD achieves significant improvements on the Flan-T5 series but is not universally applicable. First, the selection based on rationale difficulty requires the student model to have a basic capability, making it unsuitable for models without fine-tuning. Applying MoRSD to such models would require instruction fine-tuning, increasing computational costs. Second, selecting high-quality rationales requires filtering a large dataset from the teacher model, matching the computational cost of traditional CoT distillation. Future work could focus on efficient rationale generation. Moreover, the selection method relies on the student model's perplexity, which may introduce bias due to its parameter size. While small RD identifies most high-quality samples, it cannot exclude all low-quality rationales, potentially affecting distillation results.

\section*{Acknowledgments}
This work is supported by the National Science and Technology Major Program  (2024ZD01NL00101), the Major Key Project of PCL (PCL2025A03), the Natural Science Foundation of China (62506182, 62276082), National Key RD Program of China (2023YFC3502900), Shenzhen Science and Technology Research and Development Fund (KJZD20240903102802003), Shenzhen Soft Science Research Program Project (RKX20220705152815035), Shenzhen Science and Technology Research and Development Fund for Sustainable Development Project (GXWD20231128103819001, KCXFZ20201221173613036, 20230706140548006) and Guangdong Provincial Key Laboratory (2023B1212060076).

\bibliography{custom}

\clearpage
\appendix
\section{Appendix}
\label{Appendix}

\subsection{Datasets}
A summary of the datasets used in our experiments, along with their original licenses, can be found in Appendix Table \ref{datasets}. We utilize the 7 datasets from \cite{kojima2023large} to evaluate reasoning performance.

\begin{table}[ht!]
\resizebox{\linewidth}{!}{
\begin{tabular}{lccll}
\toprule
Dataset & Training Samples & Test Samples & Data Split & License  \\
\midrule
SingleEq & 356 & 152 & 70:30 & None \\
AddSub & 276 & 119 & 70:30 & Unspecified \\
MultiArith & 420 & 180 & 70:30 & Unspecified \\
SVAMP & 700 & 300 & 70:30 & MIT \\
Date Understanding & 258 & 111 & 70:30 & Apache-2.0 \\
Tracking Shuffled Objects & 525 & 225 & 70:30 & Apache-2.0 \\
StrategyQA & 1603 & 687 & 70:30 & Apache2.0 \\
\bottomrule
\end{tabular}
}
\caption{Description of datasets used in our study.}
\label{datasets}
\end{table}

\subsection{Experimental details}
All experiments were conducted on a cluster of NVIDIA V100 GPUs. We strictly controlled the hyperparameters for all datasets. For each experiment, we used a batch size of 8 and a maximum of 10,000 steps, which was found to be sufficient for the test accuracy to plateau. We report the best accuracy achieved within these 10,000 steps.

\subsection{KDE visualization of API scores}
In Section \ref{Analyse of selected rationale}, we used the ChatGPT-API to score rationales on a scale of 1 to 10 and employed KDE to visualize the score distributions for rationales selected by different methods. The KDE distributions for rationales selected via the minimum RD approach (red curves) show distinct advantages across tasks, with scores concentrated between 6 and 8, indicating higher and more consistent quality compared to other methods. The mean values of these distributions (dashed red lines) are consistently higher than those of maximum RD rationales (dashed blue lines), further supporting the superiority of the minimum RD method.

However, tasks like StrategyQA and Tracking Shuffled Objects exhibit longer tails in the minimum RD distributions, indicating a small proportion of lower-quality outliers. Despite this variability, the minimum RD method generally selects higher-quality rationales, making it a more effective approach for ensuring better overall quality in most cases.

\begin{figure}[ht!]

	\centering
        \includegraphics[width=0.5\textwidth]{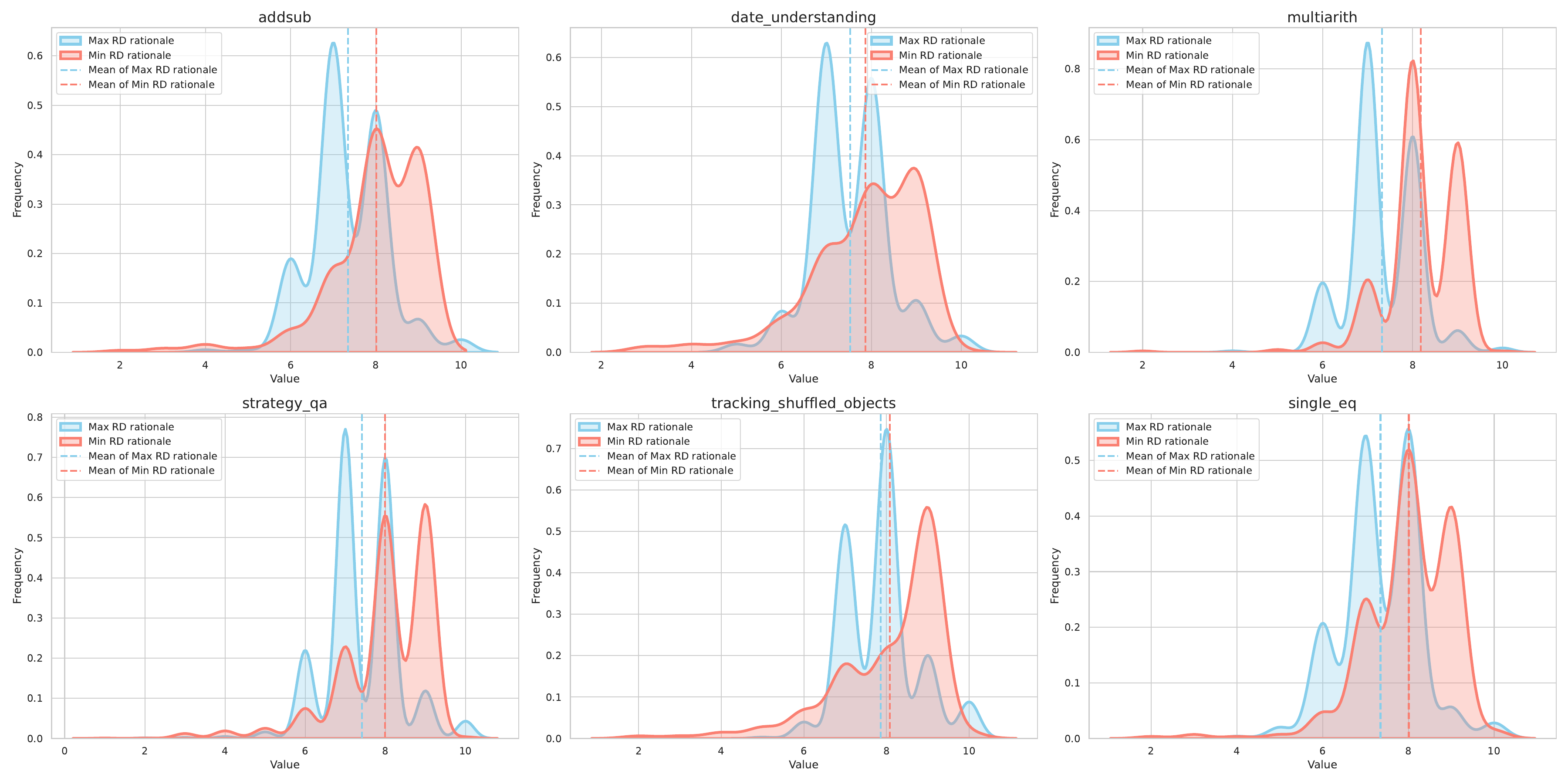} \\
	
 \caption{\textbf{KDE plot of scored selected rationale.} Kernel Density Estimation (KDE) plot, where the ChatGPT API is employed as a referee to investigate the characteristics of various rationales selected through RD. By combining different reasons and assigning them scores ranging from 1 to 10..}
\label{api_score_kde}
\end{figure}

\begin{table*}[h!]
  \centering
  \begin{tabular}{c|p{12cm}}
    \toprule
    \multicolumn{2}{l}{Prompt for Performance Evaluation} \\
    \midrule
    \multirow[c]{2}{*}{\textbf{System Prompt}} & You are a helpful and precise assistant for checking the quality of the rationale based on a given question. \\
    \midrule
    \multirow[c]{8}{*}{\textbf{Task Discribe}} &  We would like to request your feedback on the performance of two rationales in response to the question displayed above. Please rate the rationales. Each rationale receives an overall score on a scale of 1 to 10, where a higher score indicates better overall performance. Please first output a single line containing only two values indicating the scores for rationale 1 and rationale 2, respectively. The two scores are separated by a space. In the subsequent line, please provide a comprehensive explanation of your evaluation and fully compare the quality of the two rationales, avoiding any potential bias and ensuring that the order in which the rationale was presented does not affect your judgment. \\
    \midrule
    \multirow[c]{3}{*}{\textbf{Prompt}} & \texttt{[Question] \{question\} \ \ [The Start of Rationale1] \{rationale\_1\} \ [The End of Rationale1] \ [The Start of Rationale2] \{rationale\_2\} \ [The End of Rationale2] \ [System] \{TASK\_DISCRIBE\}} \\

    \bottomrule
    
  \end{tabular}
    \caption{The prompt we used to request ChatGPT to evaluate the rationales.}
    \label{prompt}
\end{table*}

\subsection{RD and length}
\begin{figure}[ht!]
    \centering
    \includegraphics[width=0.45\textwidth]{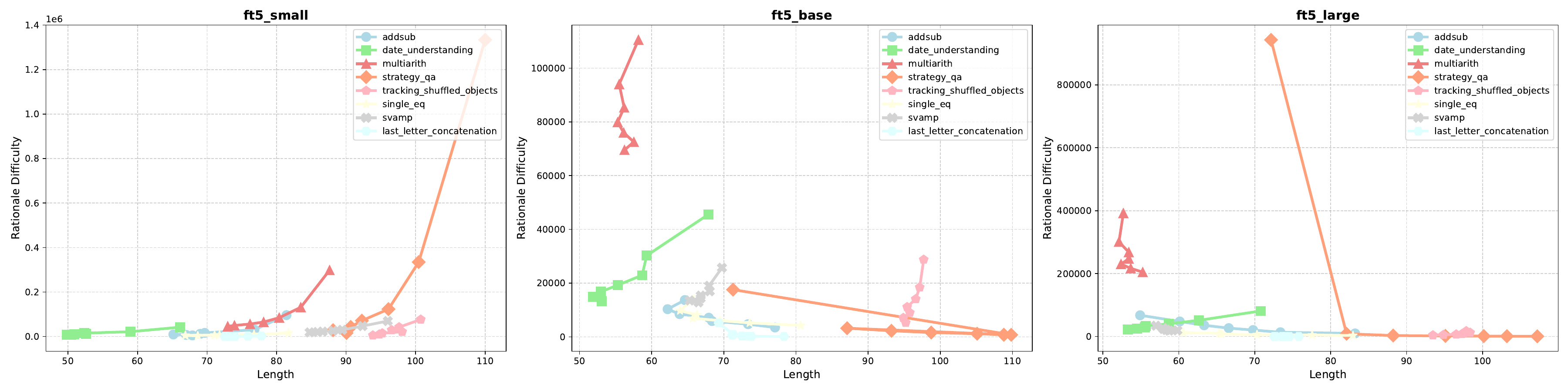} \\
    \caption{Comparison of RD Lengths}
    \label{rd_lengths_comparison}
\end{figure}

Figure \ref{rd_lengths_comparison} illustrates the relationship between rationale length and tokenized rationale length for different model sizes of Flan-T5 $\{$small, base, large$\}$. As the rationale length increases, the tokenized rationale length grows correspondingly, with a more pronounced increase observed in larger model versions. For the Flan-T5-small model, the rate of growth is moderate, indicating that smaller models require fewer tokens for shorter rationales. In contrast, the Flan-T5-base model shows a steeper increase in tokenized length as rationale length grows, reflecting its enhanced capacity to handle more complex reasoning. The Flan-T5-large model exhibits the most significant acceleration in tokenized rationale length, suggesting that larger models, with their greater capacity, demand significantly more tokens for longer rationales. This trend highlights the models' scaling behaviour, where larger models can handle more extensive rationales, necessitating an increase in the number of tokens for effective representation. Overall, the results underscore the positive correlation between rationale length and tokenized length across all model sizes, with the rate of increase being more pronounced in larger models.

\begin{table*}[h!]
\centering
\resizebox{\textwidth}{!}{
\begin{tabular}{c | c c c c | c c c c | c c c c}
\toprule
\multirow{2}{*}{\diagbox{Cal RD model}{Train model}} 
& \multicolumn{4}{c}{\textbf{AddSub}} & \multicolumn{4}{c}{\textbf{SingleEq}}  & \multicolumn{4}{c}{\textbf{StrategyQA}} \\
\cline{2-13}
\\
& \textbf{Flan-T5} & \multirow{2}{*}{\textbf{p-value}} & \textbf{Flan-T5} & \multirow{2}{*}{\textbf{p-value}} & \textbf{Flan-T5} & \multirow{2}{*}{\textbf{p-value}} & \textbf{Flan-T5} & \multirow{2}{*}{\textbf{p-value}} & \textbf{Flan-T5} & \multirow{2}{*}{\textbf{p-value}} & \textbf{Flan-T5} & \multirow{2}{*}{\textbf{p-value}} \\
& \textbf{Small} & & \textbf{Base} & & \textbf{Small} & & \textbf{Base} & & \textbf{Small} & & \textbf{Base} \\
\midrule
Flan-T5 Small  & 4.77 & -- & -1.13 & $3.8\text{e}^{-7}$ & 3.28 & -- & -0.72 & 0.14 & 52.76 & -- & -1.39 & 0.02  \\
\midrule
Flan-T5 Base & +1.04 & $1.5\text{e}^{-5}$ & 5.64 & -- & +0.22 & 0.54 & 4.72 & -- & +1.33 & $6.2\text{e}^{-4}$ & 58.39 & -- \\
\midrule
Flan-T5 Large & +1.36 & $2.3\text{e}^{-6}$ & +1.67 & $6.4\text{e}^{-9}$ & +0.16 & 0.45 & +0.19 & 0.97 & +0.58 & 0.02 & +0.55 & $6.1\text{e}^{-7}$ \\
\midrule
LLamA2-7B-hf & +1.40 & 0.001 & +1.28 & 0.141 & +0.66 & 0.062 & +0.87 & 0.012 & +0.39 & 0.223 & +0.80 & 0.082 \\
\bottomrule
\end{tabular}
}
\caption{\textbf{Transferability analyse for RD}. Flan-T5-$\{$Small, Base, Large$\}$ and LLaMA2-7B are used to calculate their RDs, which are then used to distill Flan-T5-Small. Conversely, the RD from Flan-T5-{Small, Large} and LLaMA2-7B is used to distill Flan-T5-Base.}
\label{Transferability}
\end{table*}

\subsection{Transferability of rationale selected by RD}

To verify whether the RD calculated by different models can also improve the distillation performance on other models, we use Flan-T5-{Small, Base, Large} and the larger LLamA2-7b-hf to calculate their respective RDs and use them to fine-tune the smaller Flan-T5-Small and use the RD calculated by Flan-T5-Small to fine-tune the larger Flan-T5-Base model. The RD transferability analysis and wilcoxon signed-rank test in Table \ref{Transferability} reveals that RD transfer from different models (Flan-T5 variants and LLaMA2-7B) improves performance more on simpler tasks than on complex ones. For tasks like AddSub and SingleEq, RD transfer from Flan-T5 Base and Large results in notable improvements, with Flan-T5 Large showing increases of 1.36$\%$ in AddSub (p-value = 0.009) and 1.68$\%$ in SingleEq (p-value = 0.001). However, the gains are minimal for the more complex StrategyQA task, with Flan-T5 Large only improving performance by 0.58$\%$ (p-value = 0.269). Overall, the transfer of reasoning capabilities through RD (Rationale Distillation) proves to be more effective for relatively simple tasks, where smaller models benefit significantly from the distillation process. In contrast, the impact of using larger models in such tasks tends to be less pronounced.

\subsection{Prompt for evaluation}
In this section, we provide the detailed prompt we used for evaluating the performance of two rationales for the same instruction as shown in Table \ref{prompt}

\subsection{Pattern Characteristics Comparison of Selected rationale}

In order to better compare the quality difference between the maximum RD and minimum RD rationales, we use ChatGPT’s API to compare them and give an explanation. As shown in Table \ref{Example1} \ref{Example2} and \ref{Example3}, the primary advantage of the rationale with min RD over the rationale with max RD is its more detailed and coherent reasoning process. It clearly breaks down each step of the reasoning, providing explicit explanations for how the final conclusion is reached, which enhances both transparency and logical rigor. By systematically deconstructing the problem, the rationale with min RD allows readers to more easily follow the reasoning flow. In contrast, the rationale with max RD, while more concise, may lack sufficient detail and explanation, potentially causing confusion. As a result, the rationale with min RD generally leads to a clearer understanding of the reasoning process.

\begin{figure}[ht!]

	\centering
        \includegraphics[width=0.45\textwidth]{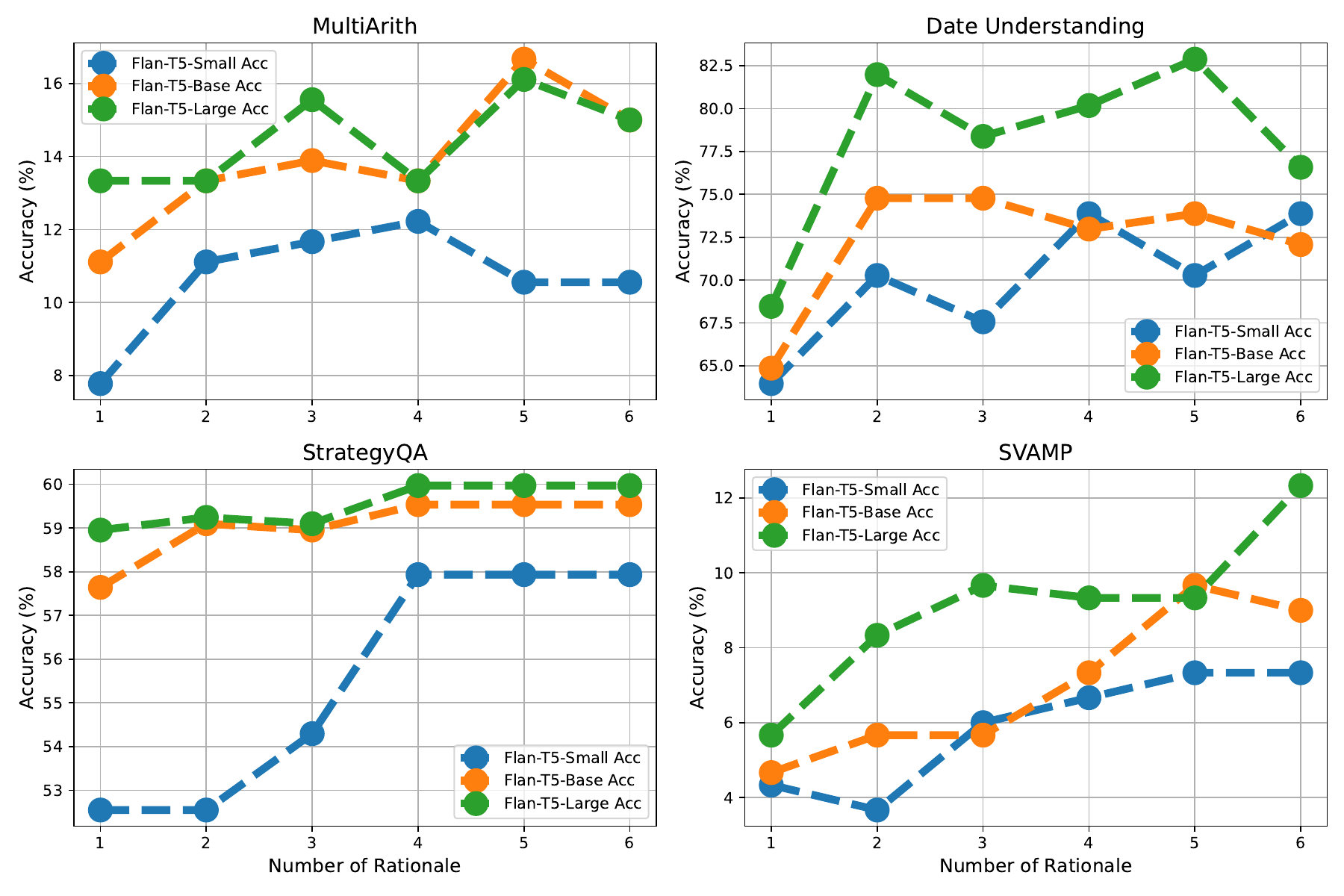} \\
	
 \caption{\textbf{Effect of the number of rationales selected by RD}. The performance of MoRSD with different numbers of rationales is evaluated on the Date Understanding, StrategyQA, SVAMP, and MultiArith datasets.}
\label{amount_analyse}
\end{figure}

\subsection{Effect of the number of rationales selected by RD}
We also examine the performance of MoRSD with varying numbers of rationales on Date Understanding, StrategyQA, Tracking Shuffled Objects, and MultiArith, respectively. The student model employed in these experiments is Flan-T5-Small. As depicted in Figure \ref{amount_analyse}, we observe that as the number of rationales increases, the model’s performance on all four datasets improves correspondingly. Specifically, when the number of rationales is increased from 1 to 4, performance is significantly enhanced on both datasets. However, the performance gains become less pronounced when the number is further increased from 4 to 6. Therefore, considering computational efficiency, we use 6 rationales in our experiments.

\begin{table*}[htbp]
  \centering
  \small
  \setlength{\tabcolsep}{6pt}
  \renewcommand{\arraystretch}{1.2}
  
  \begin{tabular}{c|>{\raggedright\arraybackslash}p{13.2cm}}
    \toprule
    \multicolumn{2}{l}{\textbf{AddSub}} \\
    \midrule
    \multirow[c]{2}{*}{\textbf{Question}} & There were 28 bales of hay in the barn . Tim stacked bales in the barn today . There are now 54 bales of hay in the barn . How many bales did he store in the barn ? \\
    \midrule
    \multirow[c]{3}{*}{\textbf{Maximum RD}} & There were 28 bales of hay in the barn. Tim stacked bales in the barn today. There are now 54 bales of hay in the barn. How many bales did he store in the barn? He must have stored 26 bales. --> 26  \\
    \midrule
    \multirow[c]{6}{*}{\textbf{Minimum RD}} & We know that there were 28 bales of hay in the barn. We know that Tim stacked bales in the barn today. We know that there are now 54 bales of hay in the barn. So, how many bales did Tim store in the barn? We can solve this problem by using basic algebra. First, we need to create a variable to represent the number of bales that Tim stored in the barn. We will use the letter "$x$" to represent this variable. Next, we need to create an equation that represents the information --> 26 \\
    \midrule
    \multirow[c]{14}{*}{\textbf{API judgement}} & 7 9 Rationale 1: Score: 7 Strengths: - Provides a direct and concise answer to the question by calculating the difference between the initial and final number of bales. Weaknesses: - Lacks explanation or reasoning behind the calculation. - Could benefit from more context or steps to show the process of arriving at the solution. Rationale 2: Score: 9 Strengths: - Demonstrates a clear step-by-step approach using algebra to solve the problem. - Provides thorough explanations and sets up an equation to represent the situation accurately. - Offers a more detailed and structured response compared to Rationale 1. Weaknesses: - Could further enhance clarity by breaking down complex steps into simpler terms for better understanding. Comparison: Rationale 2 outperforms Rationale 1 due to its comprehensive and structured approach. In conclusion, Rationale 2 offers a higher quality response by incorporating algebraic methods and providing detailed reasoning, earning it a higher score in this evaluation. \\
    \midrule
    \multicolumn{2}{l}{\textbf{Date Understanding}} \\
    \midrule
    \multirow[c]{3}{*}{\textbf{Question}} & The deadline is Jun 1, 2021, which is 2 days away from now. What is the date 10 days ago in MM/DD/YYYY? Which choice is true? Answer choices: (A) 06/24/2021, (B) 04/06/2021, (C) 04/29/2021, (D) 05/21/2021, (E) 05/20/2021. \\
    \midrule
    \multirow[c]{2}{*}{\textbf{Maximum RD}} & The deadline is Jun 1, 2021, which is 2 days away from now. So, 10 days ago would be May 21, 2021. Therefore, the answer is (D). --> E  \\
    \midrule
    \multirow[c]{2}{*}{\textbf{Minimum RD}} & Jun 1, 2021 is 2 days away from now. 2 days ago would be May 30, 2021. 10 days ago from May 30, 2021 would be May 20, 2021. Therefore, the answer is (E) 05/20/2021. --> E \\
    \midrule
    \multirow[c]{10}{*}{\textbf{API judgement}} & 7 9 Rationale 1: Score: 7 Explanation: This rationale correctly identified that the deadline is on Jun 1, 2021, which is 2 days away from the current date. The final answer choice selected was incorrect, but the reasoning process was logical and clear. Rationale 2: Score: 9 Explanation: This rationale not only correctly identified the current date and calculated 10 days ago accurately but also provided a step-by-step breakdown of the calculations. Comparison: Rationale 2 outperformed Rationale 1 due to its more detailed and coherent explanation. Rationale 2 not only provided the correct answer but also presented the steps involved in reaching that answer clearly. Rationale 1, although accurate in its calculation, lacked the detailed breakdown that Rationale 2 offered. \\

    \bottomrule
    
  \end{tabular}
    \caption{Example of Selected rationale and API judgement.}
    \label{Example1}
\end{table*}

\begin{table*}[htbp]
  \centering
  \begin{tabular}{c|p{12cm}}
    \toprule
    \multicolumn{2}{l}{\textbf{Multiarith}} \\
    \midrule
    \multirow[c]{3}{*}{\textbf{Question}} & For Halloween Debby and her sister combined the candy they received. Debby had 32 pieces of candy while her sister had 42. If they ate 35 pieces the first night, how many pieces do they have left? \\
    \midrule
    \multirow[c]{3}{*}{\textbf{Maximum RD}} & Debby took 24 pictures at the zoo, and then she took 12 more at the museum, so she took a total of 36 pictures. If she later deleted 14 pictures, she would have 36 - 14, or 22 pictures. --> 22 \\
    \midrule
    \multirow[c]{2}{*}{\textbf{Minimum RD}} & Debby took 24 pictures at the zoo and 12 at the museum. That's a total of 36 pictures. If she later deleted 14 of the pictures, she would still have 22 pictures. --> 22 \\
    \midrule
    \multirow[c]{9}{*}{\textbf{API judgement}} & 7 9 Rationale 1: The rationale correctly identifies the total number of pictures Debby took (36) and subtracts the number she deleted (14) to find the remaining amount. The explanation is clear and concise, making it easy to follow the logic. However, it could be improved by explicitly stating the final answer instead of just showing the calculation. Rationale 2: This rationale also correctly calculates the total number of pictures, mentions the deletion process, and provides the final answer. In addition to the clear logic, it includes a better structure with separate lines for the calculations, making it easier to read and understand. Overall, Rationale 2 is rated higher due to its better organization and presentation of information. While Rationale 1 is also accurate, it lacks the clarity and structure of Rationale 2. \\
    \midrule
    \multicolumn{2}{l}{\textbf{Strategy QA}} \\
    \midrule
    \textbf{Question} & Would toast for a vegan have margarine instead of butter? \\
    \midrule
    \multirow[c]{3}{*}{\textbf{Maximum RD}} & Toast is bread that has been heated until it is crisp. Bread is made from flour, water, yeast, and salt. Flour is made from wheat. So, no, toast would not have margarine instead of butter. --> Yes  \\
    \midrule
    \multirow[c]{6}{*}{\textbf{Minimum RD}} & First, what is veganism? Veganism is the practice of abstaining from the use of animal products, particularly in diet, and an associated philosophy that rejects the commodity status of animals. So, a vegan does not eat any animal products. This includes, but is not limited to, meat, poultry, fish, dairy, eggs, and honey. Now, let's think about toast. Toast is a quick and easy breakfast food that can be made with just a few ingredients. Bread and butter are the two main ingredients in toast. Bread is vegan. There are many --> Yes \\
    \midrule
    \multirow[c]{9}{*}{\textbf{API judgement}} & 6 9 Rationale 1 provides a basic explanation of veganism and the restrictions it entails in terms of avoiding animal products. It correctly points out that a vegan would not consume butter due to it being an animal product. However, the rationale lacks a direct answer to the question regarding margarine as an alternative for vegans. On the other hand, Rationale 2 concisely explains the ingredients of toast and clearly states that toast would not have margarine instead of butter. It directly addresses the question by explaining that toast consists of basic plant-based ingredients and does not require butter or margarine. Therefore, Rationale 2 scores higher for providing a more straightforward and accurate response to the given question compared to Rationale 1. \\

    \bottomrule
    
  \end{tabular}
    \caption{Example of Selected rationale and API judgement.}
    \label{Example2}
\end{table*}

\begin{table*}[htbp]
  \centering
  \small
  \begin{tabular}{c|p{12cm}}
    \toprule
    \multicolumn{2}{l}{\textbf{Single Eq}} \\
    \midrule
    \multirow[c]{2}{*}{\textbf{Question}} & Sally had 760 quarters in her bank. She spent 418 of her quarters. How many quarters does she have now ? \\
    \midrule
    \multirow[c]{3}{*}{\textbf{Maximum RD}} & Sally has 760 quarters in her bank. She spends 418 of her quarters. That means she has lost 418 quarters. So we have to subtract 418 from 760. 760-418=342 Therefore, Sally has 342 quarters left in her bank. --> 342.0 \\
    \midrule
    \multirow[c]{2}{*}{\textbf{Minimum RD}} & Sally had 760 quarters in her bank. She spent 418 of her quarters. This means that Sally now has 760 - 418 = 342 quarters in her bank. --> 342.0 \\
    \midrule
    \multirow[c]{9}{*}{\textbf{API judgement}} & 5 7 Rationale 1: The rationale correctly identifies the initial number of quarters Sally had, the amount she spent, and calculates the remaining quarters accurately. Rationale 2: This rationale also correctly identifies the initial and remaining quarters but presents the solution more concisely compared to Rationale 1. Overall, Rationale 2 is rated higher (7) due to its clarity and conciseness, while Rationale 1 is rated at 5 as it provides the correct solution but in a slightly more convoluted manner. Both solutions correctly arrive at the answer, but Rationale 2 presents the information in a more straightforward and efficient way. \\
    \midrule
    \multicolumn{2}{l}{\textbf{Tracking shuffled objects}} \\
    \midrule
    \multirow[c]{5}{*}{\textbf{Question}} & Alice, Bob, and Claire are playing a game. At the start of the game, they are each holding a ball: Alice has a black ball, Bob has a brown ball, and Claire has a blue ball. As the game progresses, pairs of players trade balls. First, Bob and Claire swap balls. Then, Alice and Bob swap balls. Finally, Claire and Bob swap balls. At the end of the game, Bob has the Which choice is true? Answer choices: (A) black ball, (B) brown ball, (C) blue ball. \\
    \midrule
    \multirow[c]{5}{*}{\textbf{Maximum RD}} & First, Bob and Claire swap balls. This means that Bob now has the blue ball, and Claire has the brown ball. Then, Alice and Bob swap balls. This means that Alice now has the brown ball, and Bob has the black ball. Finally, Claire and Bob swap balls. This means that Claire now has the black ball, and Bob has the blue ball. At the end of the game, Bob has the blue ball. Therefore, the correct answer is (C). --> B  \\
    \midrule
    \multirow[c]{4}{*}{\textbf{Minimum RD}} & output:" After Bob and Claire swap, Claire has the brown ball and Bob has the blue ball. Then, after Alice and Bob swap, Alice has the brown ball and Bob has the black ball. Finally, after Claire and Bob swap, Claire has the black ball and Bob has the brown ball. Therefore, answer choice B, Bob has the brown ball, is correct. --> B \\
    \midrule
    \multirow[c]{13}{*}{\textbf{API judgement}} & 7 9 Rationale 1 provides a clear and logical explanation of how the ball swapping between the players leads to Bob ending up with the blue ball. The rationale demonstrates a good understanding of the question and effectively leads to the correct answer. Rationale 2 also correctly follows the ball swapping sequence and provides a coherent explanation, leading to the conclusion that Bob ends up with the brown ball. The explanation is concise and clear, with proper transitions between the steps. This rationale also exhibits a good grasp of the question and leads to the correct answer smoothly. When comparing the two rationales, Rationale 2 receives a higher score due to its more concise and straightforward explanation. Additionally, Rationale 2 has a lower perplexity value, indicating more concise reasoning. However, both explanations ultimately reach the correct answer and demonstrate a solid understanding of the game's mechanics. \\

    \bottomrule
    
  \end{tabular}
    \caption{Example of Selected rationale and API judgement.}
    \label{Example3}
\end{table*}

\end{document}